\documentclass{article}

\PassOptionsToPackage{square,numbers}{natbib}

\usepackage[final]{nips_2017}

\usepackage[utf8]{inputenc} 
\usepackage[T1]{fontenc}    
\usepackage{hyperref}       
\usepackage{url}            
\usepackage{booktabs}       
\usepackage{amsfonts}       
\usepackage{nicefrac}       
\usepackage{microtype}      
\usepackage{color}
\usepackage[dvipsnames]{xcolor}

\usepackage{graphicx}
\usepackage{tabularx}

\title{Pedagogical learning}

\author{
  Long Ouyang\\
  \texttt{longouyang@post.harvard.edu} \\
  \And
  Michael C. Frank \\
  Department of Psychology \\
  Stanford University\\
  \texttt{mcfrank@stanford.edu} \\
}

\begin{document}

\maketitle

\begin{abstract}
  A common assumption in machine learning is that training data are i.i.d. samples from some distribution.
  Processes that generate i.i.d. samples are, in a sense, \emph{uninformative}---they produce data without regard to how good this data is for learning.
  By contrast, cognitive science research has shown that when people generate training data for others (i.e., teaching), they deliberately select examples that are helpful for learning.
  Because the data is more informative, learning can require less data.
  Interestingly, such examples are most effective when learners \emph{know} that the data were pedagogically generated (as opposed to randomly generated).
  We call this \emph{pedagogical learning}---when a learner assumes that evidence comes from a helpful teacher.
  In this work, we ask how pedagogical learning might work for machine learning algorithms.
  Studying this question requires understanding how people actually teach complex concepts with examples, so we conducted a behavioral study examining how people teach regular expressions using example strings.
  We found that teachers' examples contain powerful clustering structure that can greatly facilitate learning.
  We then develop a model of teaching and show a proof of concept that using this model inside of a learner can improve performance.

\end{abstract}

\section{Introduction}
{

  A common assumption in machine learning is that training data are generated by some uninformative process, such as i.i.d. sampling.
  This assumption might be justified in some cases (e.g., if we gather naturally occurring examples, like pictures of hand written digits).
  However, there are also cases where the generative process for the data is clearly informative.
  For instance, we might want to helpfully \emph{teach} a machine learning algorithm by feeding it the right examples \cite{zhu15}.
  Recent work in cognitive science \cite{shafto14} nicely demonstrates that teaching data have very different properties and behavior compared to uninformatively generated data.
  In this work, human subjects had to teach the boundary of a rectangle by labeling points that lie either within the rectangle or outside of it.
  Subjects did not simply sample points at random and label them---they deliberately selected points that would make it easy for a learner to infer the rectangle.
  In particular, they placed examples near the corners of the rectangle:

  \begin{figure}[h]
    \centering
    \includegraphics[width=0.1\linewidth]{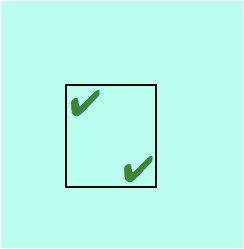}    \includegraphics[width=0.1\linewidth]{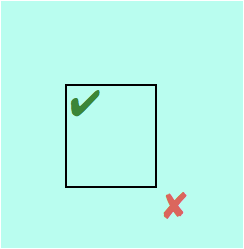} \includegraphics[width=0.1\linewidth]{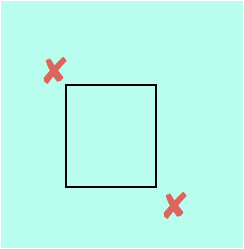}
    \caption{Helpful examples in the rectangle game.}
    \label{fig:rectangle-game}
  \end{figure}

Subsequent learners performed better when they were given these helpfully selected points.
Crucially, though, learners had to \emph{know} that points were selected helpfully---when they were given helpful examples but were told that these were generated randomly, their performance suffered.
We call this \emph{pedagogical learning}---when a learner assumes that evidence comes from a helpful teacher.
Adopting a pedagogical lens during learning appears to function as a kind of boosting---it allows learners to make stronger inferences.
While a PAC learning viewpoint would say that rectangle concepts are learnable in time polynomial to the concept size, in the pedagogical setting, it appears that we can do quite well with simply 2 examples---those that mark the corners.
There appears to be much signal in pedagogical examples.

How might we write machine learning algorithms that mine this signal?
And how much of an advantage would we gain by doing so?
Progress on these questions requires understanding how people teach complex concepts using examples.
While past work has investigated human teaching for simple concepts like rectangle boundaries \cite{shafto14} or 1D threshold rules \cite{khan11}, here we study a complex realistic setting: teaching regular expressions by giving examples of strings that either match or do not match them.
We find that helpful examples have a rich clustering structure that renders them both non-i.i.d. and non-exchangeable but these clusters can greatly facilitate learning.
Using this behavioral data, we demonstrate a proof of concept technique for adding pedagogical reasoning an existing learning algorithm and we find that the pedagogical learner outperforms its non-pedagogical counterpart.

\subsection{Related work}

Our work is inspired by cognitive science research and considers whether recent developments there might prove useful in computer science.
In particular, we draw on a computational model of teaching \cite{shafto14}, which itself is derived from models of cooperative communication \cite{frank12, golland10}.

In the realm of computer science, our work is closely related to \emph{machine teaching} \cite{zhu15}.
By and large, work in that area focuses on building machine teachers to optimize non-pedagogical learners.
Our work takes a complementary perspective---we expect that data comes from human teachers and we wish to take advantage of it by building pedagogical machine learners.

}
\section{Behavioral study: teaching}

{

  \subsection{Design and procedure}
We devised four regular expressions (regexes) and asked Mechanical Turk workers to aid in teaching the regex to a friend by generating a corpus of helpful examples of either positive strings that match the regex or negative strings that don't match the regex.
Most Mechanical Turk workers do not have experience with regexes, so we described them as rules in English (Figure~\ref{fig:teacher-ui}).

\begin{figure}[h]
  \begin{minipage}[t]{0.5\linewidth}
    \vspace{0pt}
    \includegraphics[width=0.95\linewidth]{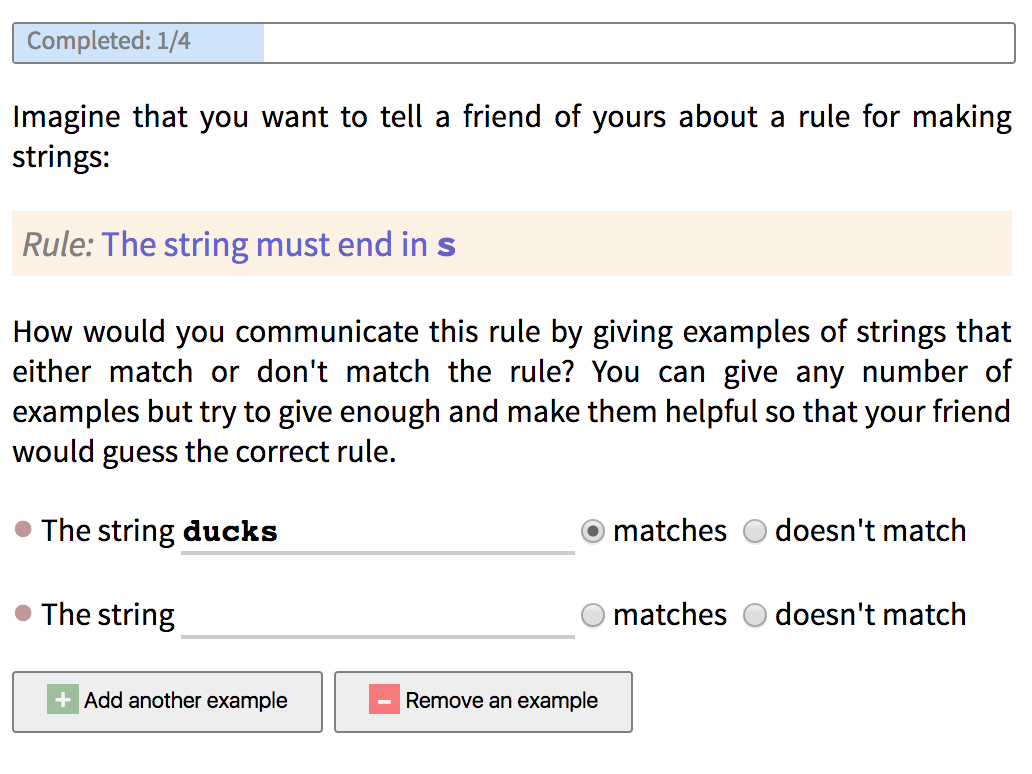}\\
  \end{minipage}
  \begin{minipage}[t]{0.5\linewidth}
    \vspace{0pt}
    {\small
      \begin{tabularx}{1.0\textwidth}[b]{l l X}
        Name & Regex & English description\\
        \hline
        3a & {\tt \string^a\{3,\}\$} & {\scriptsize The string contains \emph{only} lowercase \texttt{a}'s (no other characters are allowed) and there must be at least 3 \texttt{a}'s in the string}\\
        zip-code & {\tt {\string^}{\textbackslash}d\{5\}\$} & {\scriptsize The string is exactly 5 characters long and contains only numeric digits (0, 1, 2, 3, 4, 5, 6, 7, 8, or 9)} \\
        suffix-s & {\tt \string^.*s\$} & {\scriptsize The string must end in \texttt{s}}\\
        bracketed & {\tt {\string^}{\textbackslash}[.*{\textbackslash}]\$} & {\scriptsize The string must begin with {\tt [} and end with {\tt ]}}\\
        \hline
      \end{tabularx}}
  \end{minipage}
  \caption{Teaching user interface and regexes used. The name column is a shorthand identifier that we use to refer to the regexes throughout the paper.}
  \label{fig:teacher-ui}
\end{figure}

To assess the variability in teaching behavior, we did not constrain the number of examples (other than having to give at least 1 example), string contents, example polarities (i.e., positive vs negative), or even the correctness of the provided polarities (so to assess the error rate for this task).

We collected data from 40 subjects and discarded data from one subject who had a duplicate IP address as a previous subject.\footnote{This prevents subjects who have multiple Mechanical Turk accounts from completing the study more than once.}
The mean age of the subjects was 33.3 years old, there were 24 males and 15 females, and there was a range of experience with programming and regular experience, although most subjects had experience with neither.
Each subject generated 4 corpora --- one for each of the 4 regexes in randomized order.

\subsection{Results}

Corpora contained between 1 and 13 examples and example polarities were mostly correct (error rate of 6.6\% averaging across all 4 rules).
Corpora were fairly balanced in polarity; they contained close to equal numbers of positive and negative examples, although there was variation by rule.
For the zip-code rule, teachers gave significantly more negative examples (M = 0.64, SD = 1.85 more negative examples than positive), $t$(38) = -2.15, $p$ < 0.05 by a paired $t$-test.
For the suffix-s rule, teachers gave significantly more positive examples than negative (M = 0.69, SD = 1.21), $t$(38) = 3.55, $p$ < 0.005).

\begin{figure}[h]
  \centering
  \includegraphics[width=0.95\linewidth]{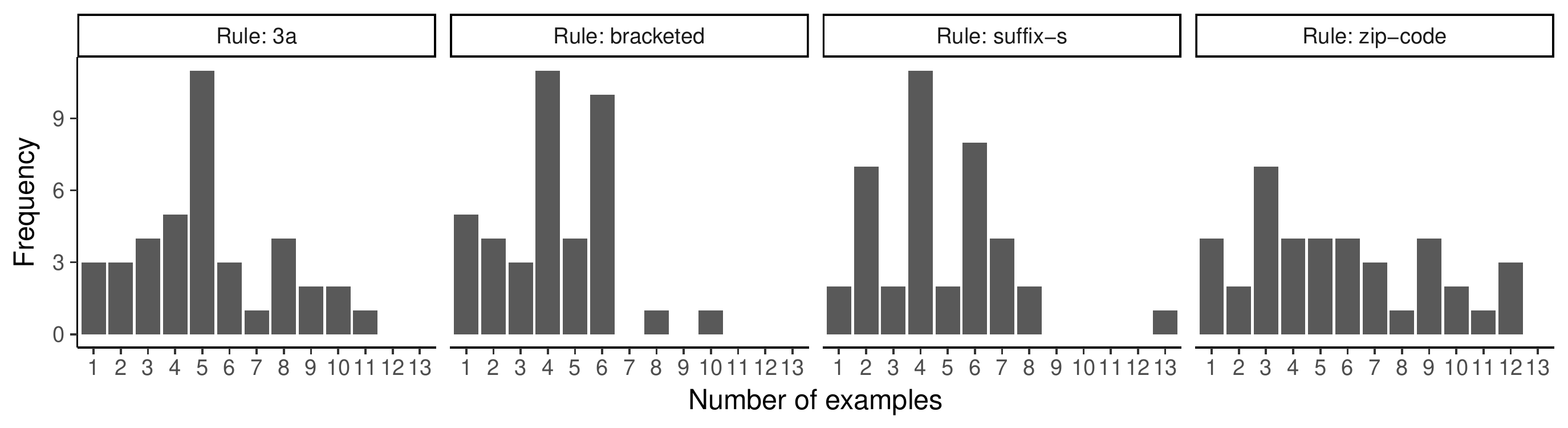}\\
  \includegraphics[width=0.95\linewidth]{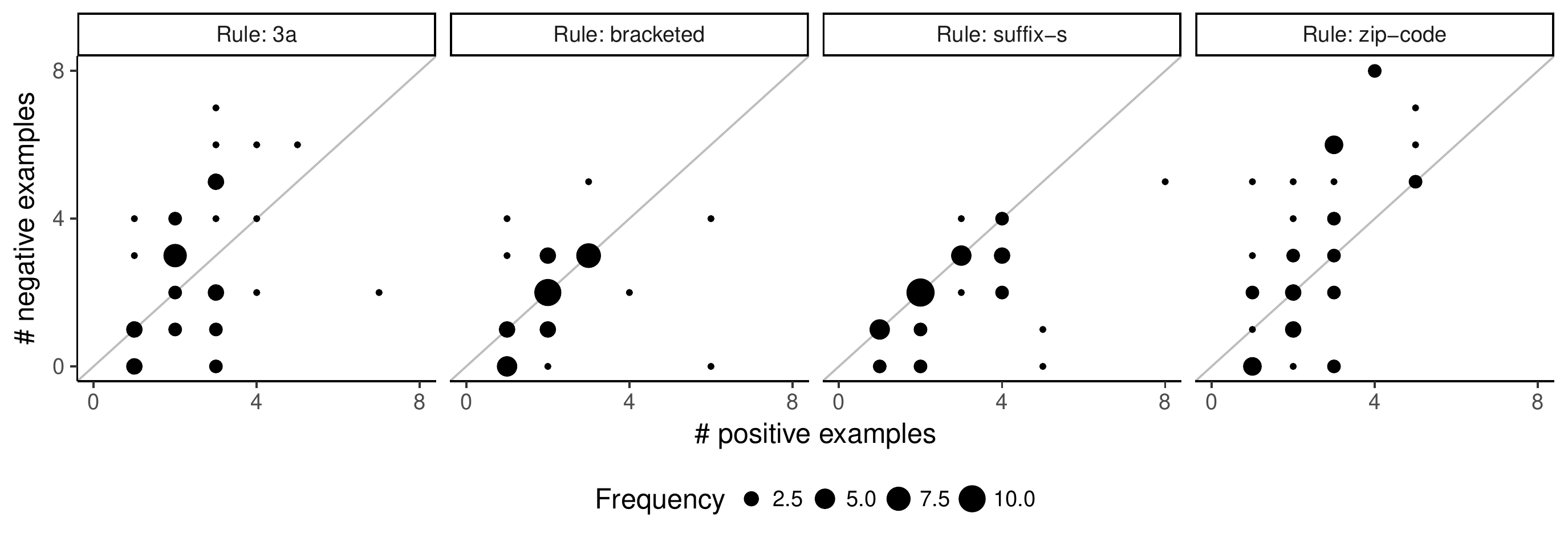}
  \caption{Corpus sizes and polarity compositions}
  \label{fig:corpora-sizes}
\end{figure}

Here we list some example corpora that teachers gave:
\begin{itemize}
\item {\color{teal} {\tt aaa}} +
\item {\color{teal} {\tt aaa}} + , {\color{teal} {\tt sss}} -- , {\color{teal} {\tt ads}} -- , {\color{teal} {\tt d2daaa}} -- , {\color{teal} {\tt 22222222222}} -- , {\color{teal} {\tt aaaa}} + , {\color{teal} {\tt aaaaaaaa}} + , {\color{teal} {\tt aaaaaaaaaaaaaaaaaa}} + , {\color{teal} {\tt a2a2a}} -- , {\color{teal} {\tt aa}} --
\item {\color{teal} {\tt 3214@}} -- , {\color{teal} {\tt 12345}} + , {\color{teal} {\tt 1234A}} -- , {\color{teal} {\tt 32146}} + , {\color{teal} {\tt 3214B}} -- , {\color{teal} {\tt 3214!}} --
\item {\color{teal} {\tt 01234}} + , {\color{teal} {\tt 00000}} + , {\color{teal} {\tt 58432}} +
\item {\color{teal} {\tt sneezes}} + , {\color{teal} {\tt breeze}} -- , {\color{teal} {\tt lots}} +
\item {\color{teal} {\tt 123ds}} + , {\color{teal} {\tt 456gs}} +
\item {\color{teal} {\tt ksuen]}} -- , {\color{teal} {\tt [ABC123]}} + , {\color{teal} {\tt [sjf6s]}} + , {\color{teal} {\tt [skfhme5}} -- , {\color{teal} {\tt [12345]}} + , {\color{teal} {\tt 485jdns}} --
\item {\color{teal} {\tt [dog]}} + , {\color{teal} {\tt dog}} -- , {\color{teal} {\tt [cat]}} + , {\color{teal} {\tt cat}} -- , {\color{teal} {\tt [123] +}} , {\color{teal} {\tt 123}} --
\end{itemize}
In general, the corpora have several properties that intuitively seem to support learners.
First, analogously to the patterns of teaching found in \cite{shafto14}, teachers tended to select negative examples that were close to the concept boundary---they were \emph{almost} positive examples, e.g., {\tt [re7a8} for the  bracketed rule.
This is similar to the near miss phenomenon discussed in \cite{winston70}.

Furthermore, negative examples often appeared to be constructed by taking positive examples and lightly editing them to be negative examples, e.g., for the {\tt zip-code} rule, the positive string {\tt 12345} occurring first and the negative string {\tt 1234} occurring second.
Indeed, if we cluster the strings within corpora by edit distance with a threshold of 2, we find that the number of clusters per corpus (3.27) is far lower than would be expected by chance, $p$ < 0.0001, 95\% CI for bootstrapped permutation test with 1000 samples: 4.98---5.11.
Taking these two results together, strings within corpora tend to be related to each other---positive strings tend to be minimally edited to negative strings so as to demonstrate the concept boundary.
This ``minimal pair'' phenomenon is additionally interesting because it does not appear in the simpler setting of learning a rectangle boundary \cite{shafto14}.
There, negative examples were placed close to concept boundary but not close to positive examples.
This suggests that teaching complex concepts may have different characteristics than teaching simple ones.

Finally, corpora had internal temporal structure.
The first example in a cluster tended to be positive (102 clusters) rather than negative (28 clusters), $\chi^2$(1) = 42.12, p $<$ 0.0001.
Additionally, clusters tended to occur in contiguous spans rather than broken up across the corpus sequence.

In sum, teachers appear to generate temporally ordered clusters of related examples.
These relations appear to highlight the concept boundary by contrasting a positive example with a nearby negative example.
It is worth emphasizing that this setting is quite exotic from a traditional machine learning perspective---these data are both non-i.i.d. and non-exchangeable.
Nevertheless, they do intuitively seem suited to support learners.

We verified that these examples are helpful by presenting them to human learners.
We took each corpus and presented it 10 learners and asked them to guess the rule.
We then coded each response as correct or incorrect.\footnote{This revealed that people entertain a number of complex hypotheses that are not easily captured by regular expressions (e.g., for zip-code, some thought that the numbers had to be arranged in increasing sequence and for 3a, some thought that it could be any string consisting of entirely the same letter).}
Results are shown in Figure~\ref{fig:learning-results}.
There is variation between rules and corpora (e.g., some subjects in the teaching experiment were lazy and gave only  single example for each rule), but overall, it appears that people are capable of providing good evidence about complex rules using a handful of well chosen examples.
This is particularly impressive given that we did not monetarily incentivize helpful examples.

\begin{figure}[t]
  \centering
  \includegraphics[width=0.95\linewidth]{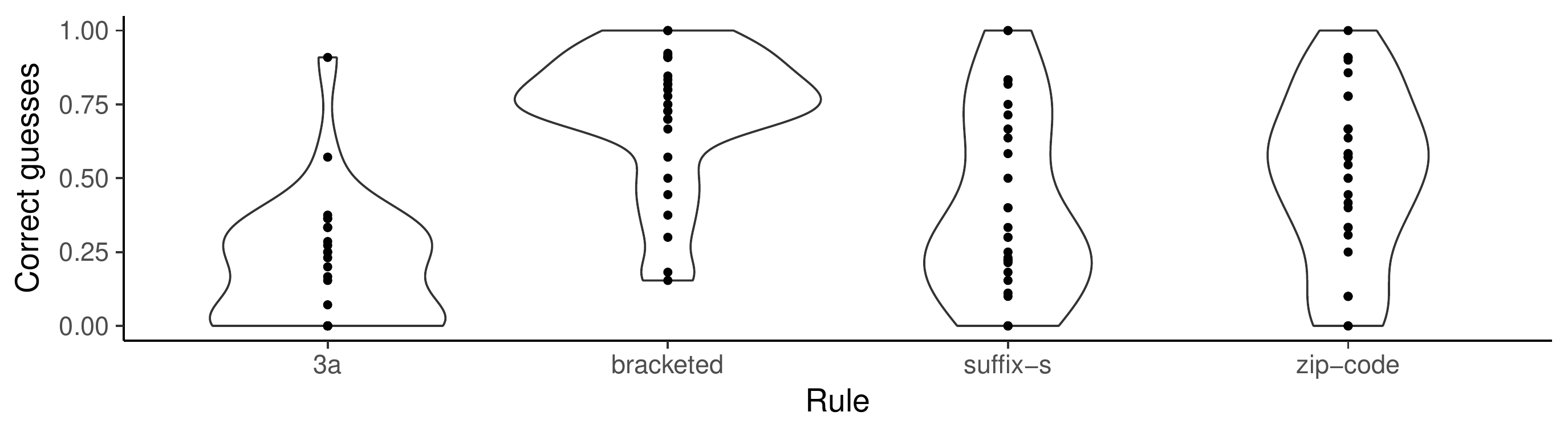}\\
  \caption{Human learning results. Each point represents a corpus.}
  \label{fig:learning-results}
\end{figure}

\section{Proof of concept: building a pedagogical learner}
{

  \subsection{Models}
  In this section, we discuss how to build a pedagogical learner to take advantage of the helpful teaching data we collected.
  This is based on a model of pedagogy from \cite{shafto14}.
  We begin by building a Bayesian non-pedagogical learner, $L_0$.
  We implement a model of a teacher, $T_1$, which aims to facilitate learning for $L_0$.
  We then implement a pedagogical learner, $L_1$, which expects to receive examples from $T_1$.
  Finally, we compare $L_0$ and $L_1$ in a synthetic experiment and show that $L_1$ outperforms $L_0$ on helpful data.

  \subsubsection{Non-pedagogical learner}

  The non-pedagogical learner $L_0$, receives a corpus $c$ of examples $\{(x_1, y_1), \dots, (x_n, y_n)\}$, where $x_i$ are strings and $y_i \in \{0, 1\}$ are labels (0 is negative, 1 is positive).
  The learner then computes a posterior on regexes $P_{L_0}(r \mid c)$, which is given by Bayes Rule,

  $$P_{L_0}(r \mid c) \propto \pi_{L_0}(r)\mathcal{L}_{L_0}(c \mid r)$$

  Here, $\pi_{L_0}(r)$ is the prior probability of $r$ and $\mathcal{L}_{L_0}(c \mid r)$ is the likelihood of observing corpus $c$ given regex $r$.
    We use a description length prior, $\pi_{L_0}(r) \propto \exp(-|r|)$, and a likelihood that softly conditions on the string-polarity pairs in $c$ being consistent with the regex $r$.
    Let $r(x)$ denote the polarity of a string $x$ according to $r$ (1 if $x$ matches $r$ and 0 if not), and let $Q_r(c)$ be the number of incorrectly labeled examples according to $r$, $Q_r(c) = \sum_{(x_i, y_i) \in c}r(x_i) \oplus y_i$, where $\oplus$ is exclusive disjunction.
    Then we set $\mathcal{L}_{L_0}(c \mid r) \propto \exp -\beta Q_r(c)$ where $\beta \in [0,1] $ is a parameter controlling tolerance for errors; at $\beta = 1$, we reject any corpora with inconsistent examples and at $\beta = 0$, all errors are ignored.
    This likelihood implicitly assumes that the data were generated by an uninformative generative process called \emph{weak sampling} in which strings are sampled i.i.d. and labeled according to $r$.

  \subsubsection{Teaching model}

  Whereas learners receive corpora and return a distribution on regexes, the teacher receives a regex to teach and returns a distribution on corpora.
  Our teaching model $T_1$ samples corpora according to
  $$P_{T_1}(c ; r) \propto [\pi_{T_1}(c)\mathcal{L}_{T_1}(r \mid c)]^\alpha$$
  where $\alpha \geq 1$ is a temperature parameter.
  The prior here represents a cost preference---all else being equal, we prefer to give fewer examples and also shorter examples:

  $$ \pi_{T_1}(c) \propto \frac{1}{2^{|c|}} \prod_{(x_i,y_i) \in c} \frac{1}{2^{|x_i|}}\delta_{r(x_i),y_i} $$

  Viewed as a generative model, the prior samples the corpus size $|c|$ from a geometric distribution, samples strings $x_i$ according to a geometric distribution on length, and  labels each example according to the regex $r$.
  Observe that this is prior is just an instance of weak sampling, which is uninformative.
  However, the likelihood encourages informativity---the teacher conditions on the $L_0$ learner correctly recovering the regex $r$ from the corpus: $\mathcal{L}_{T_1}(r \mid c) = P_{L_0}(r \mid c)$.

\subsubsection{Pedagogical learner}

The pedagogical learner uses the same prior as the non-pedagogical learner but a different likelihood.
Instead of assuming weak sampling, the pedagogical learner thinks about what corpus $c$ a helpful teacher would be likely to give as evidence if the true rule were $r$.
In other words, the likelihood is just $P_{T_1}(c ; r)$:

$$P_{L_1}(r \mid c) \propto \pi_{L_0}(r)P_{T_1}(c ; r)$$

The subscript notation on $L_0$, $T_1$, and $L_1$ is suggestive---we might imagine even more levels of mutual reasoning.
This is certainly possible but for our proof of concept we will stop here.

\subsection{Comparison}

We compared the ability of $L_0$ and $L_1$ to learn the correct regex from the teaching data we collected in the behavioral study.
Doing full inference in the learners and teachers is quite challenging, so we restricted the hypothesis spaces for both.
Computing $L_0$ over the set of all regular expressions amounts to a problem known as grammar induction, which is an active area of research.
So we restricted the space of regexes; for each of the 4 regexes from the teaching experiment, we devised 2 distractor regexes that were similar but different:

\begin{table}[h]
  \centering
  \caption{Space of regexes used in learner comparison}
  \label{tab:distractors}

  \begin{tabular}{l l l l }
    Name & Target regex & Distractor 1 & Distractor 2\\
    \hline
    3a  & {\tt \string^a\{3,\}\$ } & {\tt \string^a\{6,\}\$} & {\tt \string^[aA]+\$} \\
    zip-code  & {\tt \string^{\textbackslash}d\{5\}\$} & {\tt \string^.\{5\}\$} & {\tt \string^{\textbackslash}d+\$}\\
    suffix-s  & {\tt \string^.*s\$} & {\tt \string^.*s.*\$} & {\tt \string^.*[a-z].*\$} \\
    bracketed  & {\tt \string^{\textbackslash}[.*{\textbackslash}]\$} & {\tt \string^{\textbackslash}[.*\$} & {\tt \string^.*{\textbackslash}]\$} \\
    \hline

  \end{tabular}
\end{table}

Fully computing $T_1$ would also be difficult, as we would need to do inference over the very large space of corpora that human teachers might plausibly give.
Thus, we approximated the space of corpora by using empirical data from human teachers.
We already had this data for the target regexes from our earlier teaching study, so we only collected new teaching data for the distractors.

For each corpus that was generated for a target regex, we computed the probability of the correct answer under $L_0$ and $L_1$ for a range of values for parameters $\alpha$ and $\beta$.
As Figure~\ref{fig:synth-results} shows, $L_1$ generally outperforms $L_0$, sometimes by a dramatic margin (e.g., $L_0$ accuracy = 0.25 versus $L_1$ accuracy = 0.8).
At its worst, $L_1$ performs on par with $L_0$; these are the $\log \beta = -0.1$ column and $\alpha = 1$ row of the figure. When $\log \beta = -0.1$, the $L_0$ model is extremely tolerant of errors in its likelihood function, so little belief updating actually occurs, which makes it less suprising that $L_1$ (which depends on $L_0$ as a subroutine) would not do much better.
It is interesting that for $\alpha = 1$, $L_0$ and $L_1$ are on par---it may be that human teachers use a higher $\alpha$ value than 1.
To summarize, $L_1$ appears to outperform $L_0$, broadly speaking.
In a sense, this should not be so suprising---$L_1$ uses a more accurate model of how the data are generated.

\begin{figure}[t]
  \centering
  \includegraphics[width=1.0\linewidth]{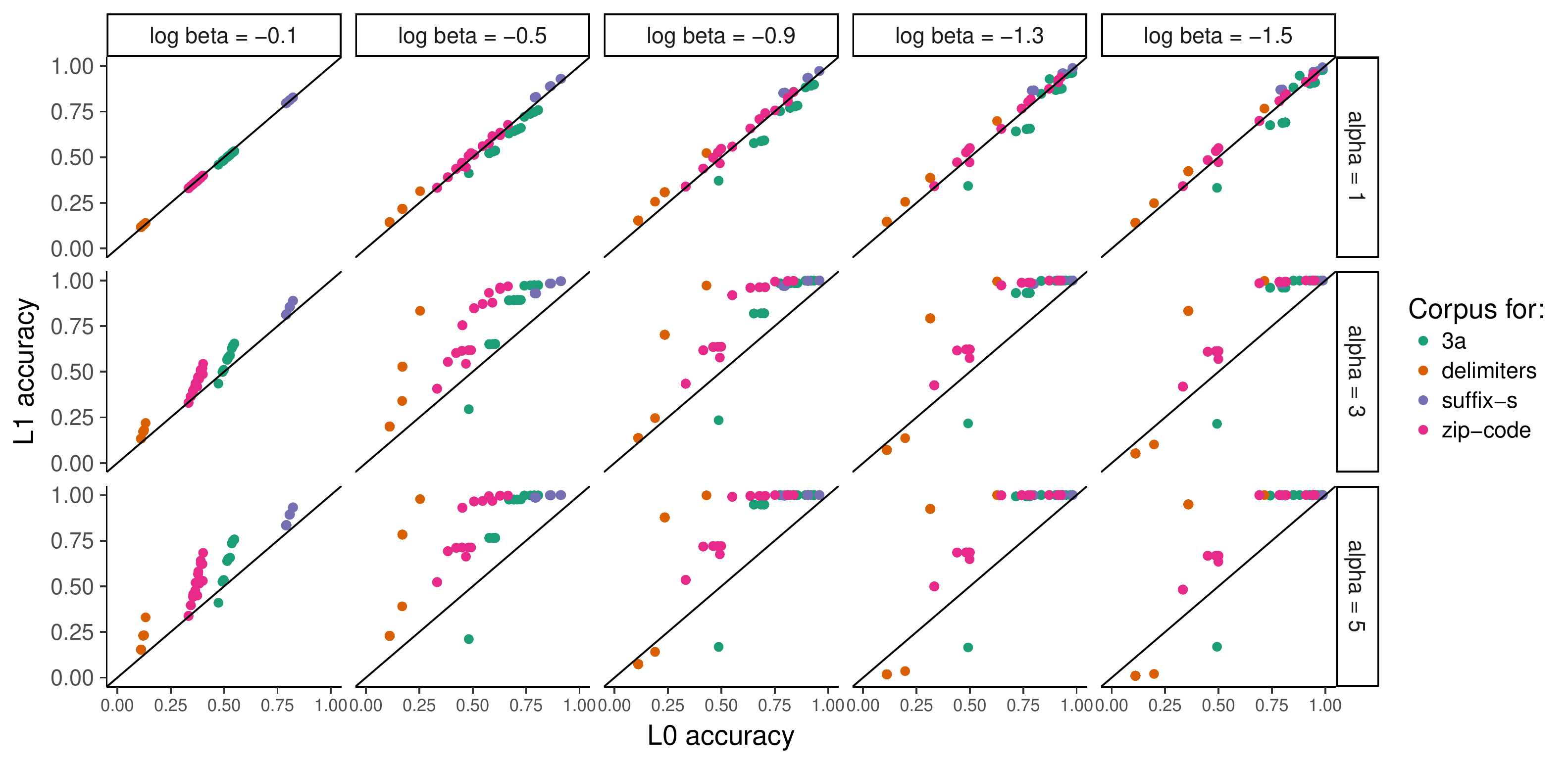}
  \caption{Comparison of non-pedagogical and pedagogical learners}
  \label{fig:synth-results}
\end{figure}

}

\section{Discussion}

In this work, we have shown that human teachers can very effectively generate examples to communicate complex concepts to learners.
In principle, then, teaching data are powerful source of evidence.
However, because current algorithms typically assume an uninformative model of the data, they are not fully equipped to take advantage of teaching---they are not pedagogical learners.
We have given a sketch of how to promote certain non-pedagogical learners into pedagogical learners: one can collect data on how human teachers generate examples for a domain and use a computational model of this process as the likelihood in Bayesian inference.

Making pedagogical machine learning useful in practice will require work in two directions.
First, we need more cognitive theory---there remains much to understand about how people teach examples for various domains.
In our work, we studied teaching of binary classifiers over a complex discrete domain and found that teachers generated helpful clusters of related examples, but it seems certain that other domains will exhibit different teaching patterns.
We hazard three conjectures that may be interesting topics for future work.
First, the near miss and minimal pair teaching patterns might not be possible for regression problems.
Second, teaching might differ when human teachers are not capable of generating examples but instead can only select or modify existing ones (e.g., identifying tumors from X-ray scans).
Third, errors in teaching examples might be particularly damaging (if learners expect teachers to be helpful, then giving mislabeled examples could cause problems).
We found low but nonzero error rates in teaching by nonexperts--- further analyzing errors may help make pedagogical learners more robust.

The second direction for future work is inference machinery.
Pedagogical learning is computationally challenging---as we have framed it, the teaching model must repeatedly call the non-pedagogical learner as a subroutine.
In turn, the teaching model is repeatedly called as a subroutine by the pedagogical learner.
And of course, things get worse if we do additional levels of reasoning.
It may be currently possible to perform pedagogical learning for relatively simple but fast learners (e.g., graphical models with low treewidth), but adding pedagogy to more complex models will require new techniques (e.g., eaves-chapter).
Additionally, it is possible that non-Bayesian versions of pedagogical learning (using, say, neural networks) could be more efficient.

\subsubsection*{Acknowledgments}

This work was supported as part of the Future of Life Institute (futureoflife.org) FLI-RFP-AI1 program, grant number 2017-175569.


\bibliographystyle{ieeetr}
\bibliography{refs}

\end{document}